\documentclass[10pt,twocolumn,letterpaper]{article}

\usepackage{cvpr}
\usepackage{times}
\usepackage{epsfig}
\usepackage{graphicx}
\usepackage{amsmath}
\usepackage{amssymb}
\usepackage{mathtools}
\usepackage{gensymb}
\DeclarePairedDelimiter{\ceil}{\lceil}{\rceil}

\usepackage{tabu}
\usepackage{booktabs}
\usepackage{times}
\usepackage{graphicx} 
\usepackage{tabularx}
\usepackage{subfigure} 
\usepackage{amsmath}

\usepackage[breaklinks=true,bookmarks=false]{hyperref}

\cvprfinalcopy 

\def\cvprPaperID{****} 

\begin{document}

\title{Conversational Group Detection With Deep Convolutional Networks}

\author{Mason Swofford\thanks{These authors contributed equally.}\\
Stanford University\\
Stanford, CA\\
{\tt\small mswoff@stanford.edu}
\and
John Peruzzi\footnotemark[1]\\
Stanford University\\
Stanford, CA\\
{\tt\small jperuzzi@stanford.edu}
\and
Marynel V\'azquez\\
Stanford University\\
Stanford, CA\\
{\tt\small marynelv@stanford.edu}
}

\maketitle

\begin{abstract}
   Detection of interacting and conversational groups from images has applications in video surveillance and social robotics. In this paper we build on prior attempts to find conversational groups by detection of social gathering spaces called o-spaces used to assign people to groups. As our contributions to the task, we are the first paper to incorporate features extracted from the room layout image, and the first to incorporate a deep network to generate an image representation of the proposed o-spaces. Specifically, this novel network builds on the PointNet architecture which allows unordered inputs of variable sizes. We present accuracies which demonstrate the ability to rival and sometimes outperform the best models, but due to a data imbalance issue we do not yet outperform existing models in our test results.
\end{abstract}

\section{Introduction}

\subsection{Motivation}

Conversational group detection from images has applications in video surveillance and social robotics. Effective detection of groups allows a mobile robot to navigate around people without interfering inappropriately with social interactions. Sociologists define a facing formation, or f-formation, as a socio-spatial formation in which people maintain a convex space termed an o-space $\cite{setti2015f}$. Detection of o-spaces and assigning people to these o-spaces is a standard and sociologically consistent method for detection of conversational groups. Robots which operate in public spaces, particularly those which interact with and serve humans, can perform their functions more naturally and seamlessly if able to recognize these social formations and spaces.

\subsection{Literature Review}

There is ample literature on group detection which sets standards for error analysis, group definition, and a baseline for performance.
The best performance to date has come not from a learning algorithm but rather a graph-cuts clustering algorithm, which attained precisions from approximately .84 to .65 on the Cocktail Party dataset depending on the threshold for accuracy used $\cite{setti2015f}$ (see Evaluation Metrics for explanation of error analysis). We will attempt to improve upon these results. 
Other attempts have included edge-weighted graph algorithms, for instance one in which each node is a person and the edge measures affinity between pairs $\cite{hung2011detecting}$. There is a similar method with weightings based on an attention metric, differing from the previous method by exploiting social cues to determine edge weightings $\cite{tran2013social}$. There is also a game theoretic framework with probabilistic overlapping regions determined by an attention metric which uses temporal data $\cite{vascon2014game}$. Finally, another successful algorithm has been a voting-based algorithm where each individual gets a vote for the o-space given by a Gaussian function for that individual $\cite{setti2013multi}$. All attempts have performed worse or similar to the graph-cuts algorithm, so we will focus on improving upon those results. Additionally, it is promising that we are the first paper to apply deep learning to this task and the first paper to use features from a layout image of the room.

\subsection{Methodology Summary}

\begin{figure}[h!]
  
  \centering
  \includegraphics[width=0.2\textwidth]{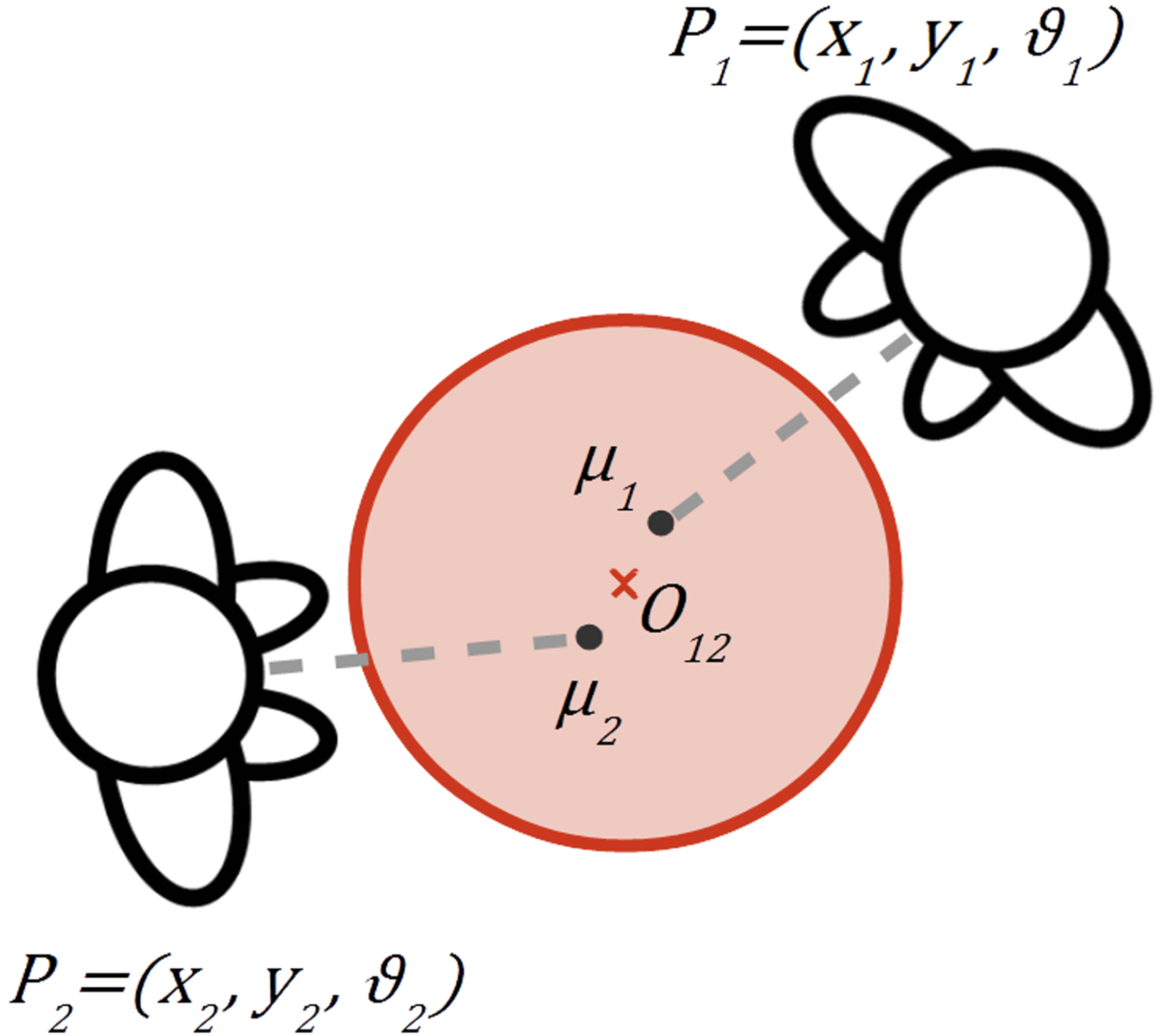}
  \caption{Example O-Space \cite{setti2015f}}
\end{figure}

Our data comes from standard datasets for f-formation detection, which are videos of social interactions in confined spaces. These are annotated by sociologists with position, orientation, and group assignment for every person in various frames taken throughout the video. We use 2D coordinates and yaw orientations of the people as features. Additionally, we manually transform the  camera images into a 2D map of the room, and extract features from this map through deep convolutions.
	
	Using the annotated features and image map, we run a deep network to output a 2D image of the room, with each pixel of the output image representing the likelihood of that location containing an o-space. We then perform non-maximal suppression and thresholding to arrive at our final predictions for the o-space locations and greedily assign people to the nearest o-space to determine our conversational groups.


\section{Data Acquisition and Manipulation}

\subsection{Datasets}

There exist multiple standard annotated datasets for the task of f-formation detection. We elected to use the Cocktail Party dataset $\cite{setti2013multi}$ due to its clarity and known accurate annotations. There are four camera angles which record a 30 minute video of a 30 sq m room, in which 6 people wander around and occasionally converse. One frame every three seconds is annotated with $x,y$ position and yaw orientation of each person, as well as which, if any, conversational group they belong to. There are a total of 320 annotated frames which we have divided with an 80\%, 10\%, 10\% split into train, validation, and testing sets, respectively. This division was not done randomly as is common practice in most machine learning applications, but instead it was done sequentially to prevent test groups and positions from resembling training groups and positions, with the validation set acting as an additional buffer between them.
\begin{figure}[h!]
  
  \centering
  \includegraphics[width=0.3\textwidth]{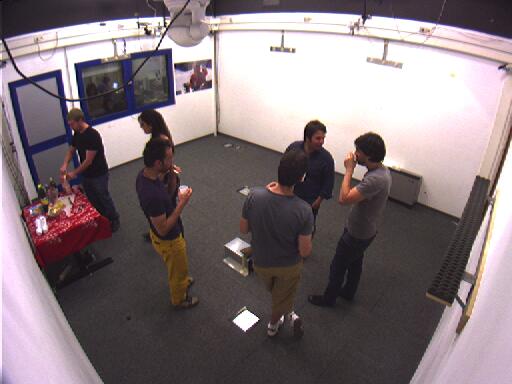}
  \caption{Example Frame from Camera \cite{setti2013multi}}
\end{figure}
	Our data therefore has both images and a file of annotated features. For future work, we plan to incorporate other standard datasets such as the Coffee-Break $\cite{setti2015f}$ or SALSA $\cite{alameda2016salsa}$ datasets to acquire more training data.

\subsection{Feature Extraction}

Position in $x,y$ coordinates of all people is left as annotated from the dataset. For head orientation, which was previously given in radians, we bucketed into 16 possible discrete orientations of size 22.5$\degree$. This is done in order to allow the network to learn that degrees wrap around. For example, a value of 359$\degree$ would be more similar to 0$\degree$ than it would to 300 $\degree$, and using continuous numerical values would not easily allow the network to learn this trait. As well, modern networks to detect head orientation algorithms often output yaw in discrete buckets rather than continuous values for similar reasons $\cite{maji2011action}$.
	
	We manually annotate the room image to create a 2D map of important features of the room, with the hope that knowledge of locations of items such as walls or tables will allow our algorithm to better identify social spaces. Transformation of pixels from frames taken by cameras to overhead 2D coordinates involves using a 3D to 2D calibration algorithm provided by the dataset to convert an $x, y, z$ point in the room to a pixel from the camera image. We set $z = 0$ as we only desire $x, y$ coordinates and then visualize $x,y,0$ points on the camera image to locate the room items in the $x,y$ plane. Modern 3D segmentation and classification algorithms will hopefully automate this process in the future.
    
\begin{figure}[h!]
	
  \centering
  \includegraphics[width=0.2\textwidth]{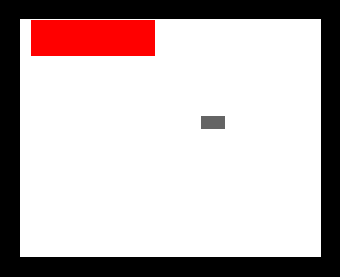}
  \centering
  \caption{Overhead Room Image}

\end{figure}

\subsection{Ground Truth from Labeled Data}

As the number and size of groups is variable and therefore difficult to output or even represent in a vectorized form, we instead have our algorithm output a 10x12 2D representation of the room, with each pixel of our 2D image corresponding to a 0.5x0.5 meter space on the floor and containing a value representing its likelihood of containing an o-space. To generate the true image for our labeled data, we find the true o-spaces using a known algorithm which minimizes distance of the o-spaces from every person in a group defined by an o-space $\cite{vazquez2015parallel}$. We label the pixels by treating the true o-spaces as the means of multi-dimensional Gaussians. To output actual groups, we assign people greedily to the o-spaces determined by our algorithm.

\begin{figure}[h!]  
  \centering
  \includegraphics[width=0.3\textwidth]{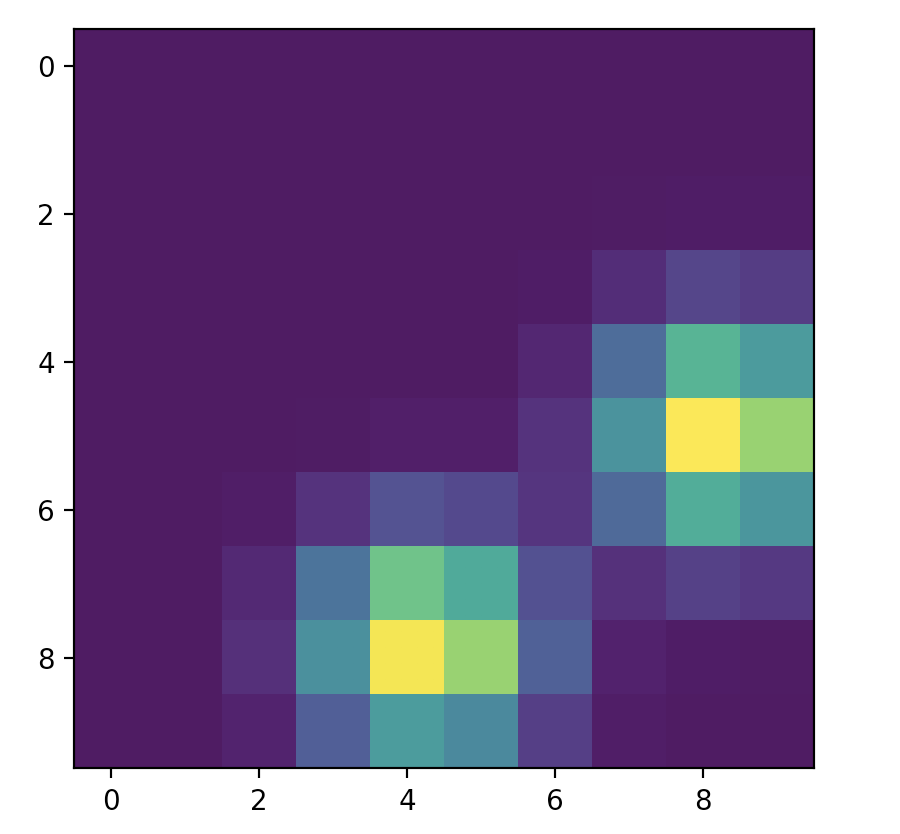}
  \caption{Ground Truth Image of 2 O-Spaces}
\end{figure}


\subsection{Data Augmentation}

Our model is very complex and the dataset consisted of only 320 annotated frames. We thus augmented the data by flipping the room across the vertical axis, horizontal axis, and both. To do this, we flipped the location and angle of each person accordingly, leaving group annotations the same. We also flipped the overhead room image before extracting its features. Through this, we increased the size of our dataset to 1280 frames. Before this augmentation, our model had difficulty learning, but afterward our accuracies in many cases doubled or more.

\section{Learning Methods}

Our model can be broken down into three connected components: room feature extraction, people feature extraction, and a fully connected network.

\subsection{Extracting Room Features}

For each room image, we wish to extract relevant features for our model. In order to do so, we feed the images into MobileNet and extract the output from block 9, approximately 3/4 of the way through the network. We theorize that this layer is deep enough to contain useful features, but not to deep as to be overly specialized for the image classification task it was trained on. We chose MobileNet because it has the ability to reduce the input size as the image progresses through the network. To do this, we chose an alpha=.25 which reduces the image by 1/4 as it passes through the network. However, the output image is still too high dimensional for our purposes, given our limited data. Therefore, we chose to perform PCA on the outputs to reduce the image to 1024 dimensions. While this process is part of our deep network, we freeze all the weights in MobileNet, as we do not have enough data to retrain it.

\subsection{Extracting People Features}

Each person is represented as an 18-dimensional vector consisting of the x coordinate, y coordinate, and 16-dimensional one hot yaw, where the x and y coordinates are normalized across the training set. However, since we do not want the order in which we feed the people into our network to matter, and we want to allow different numbers of people to be input into the network, since different datasets have different numbers of people, we use an innovative approach first used in PointNet $\cite{qi2017pointnet}$. 

We first create a 3D tensor of dimension 1xMx18 where M is the maximum number of people possible (M=25 in our case) and fill the first 1xPx18 values with our people vectors, where P is the number of people in a particular example (P=6 for the Cocktail Party dataset), and leave the remaining 1x(M-P)x18 values zero. Then, we apply a series of 1x1 convolutions to the tensor to slowly grow it to a 1xMxD tensor using a deep network, where D is our output depth, which was determined by cross validation to be 1024. Note that in this process the ordering of the person vectors  is irrelevant because each vector is growing without interacting with the other person vectors. Finally, a symmetric function, in our case the max function, is applied across the final dimension to obtain a D-dimensional vector which encodes information about all of the input people and is independent of input order. Note that unlike the room image feature extraction component, the weights in this layer are trained in response to the output loss.

\subsection{Generation of Groups}

The final component of our neural network is a series of fully connected layers that takes in the flattened MobileNet room features and PointNet people features and outputs a 120-dimensional vector which corresponds to a flattened version of our ground truth o-space map. To train our network, we evaluate our output against the ground truth using the mean squared error loss. To get our predicted groups, we convert our 120-dimensional vector to a 10x12 image and perform non-maximal suppression and thresholding to obtain potential o-spaces. Then, each person proposes an o-space given his position, orientation, and a learned stride-length. We assign each person to the potential o-spaces proposed by the non-maximal suppression neared to that person's proposed o-space. People assigned to the same o-space are then in the same group.

\begin{figure}[h!]  
  \centering
  \includegraphics[width=0.4\textwidth]{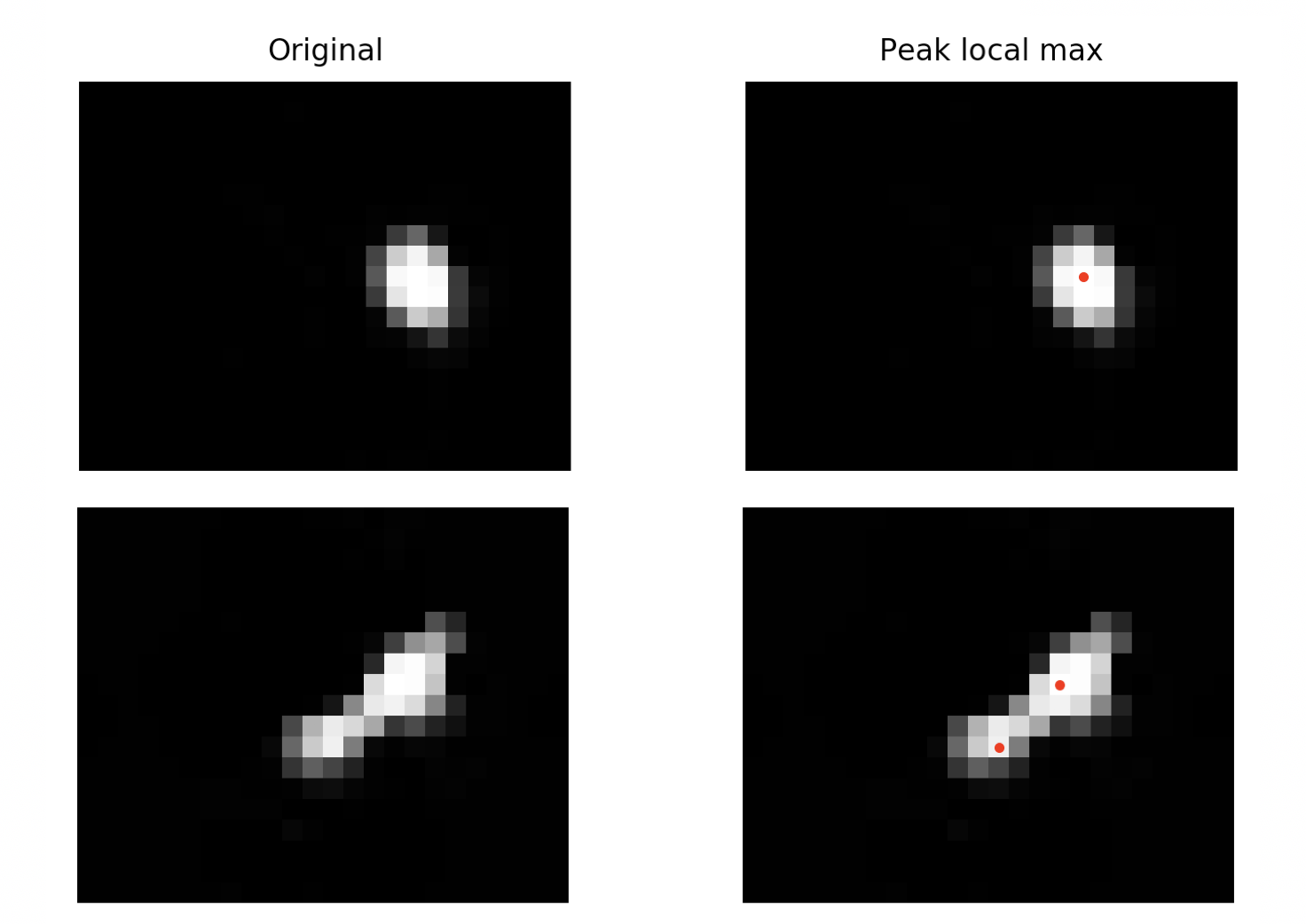}
  \centering
  \caption{Maximal non-suppression to generate potential o-space centers}
\end{figure}

\begin{figure}[h!]  
  \centering
  \includegraphics[width=0.4\textwidth]{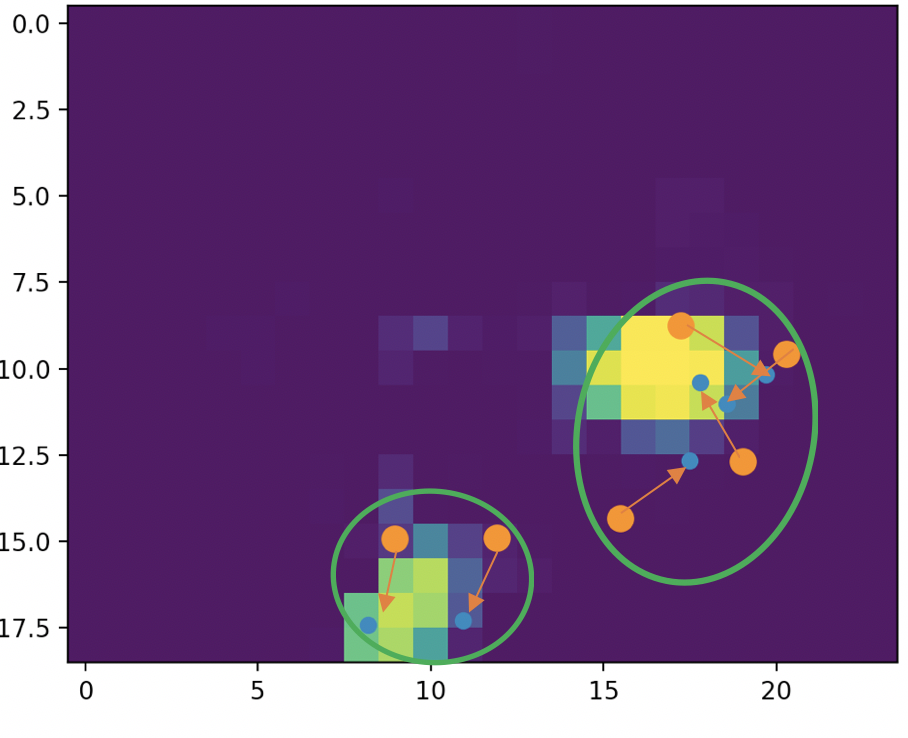}
  \centering
  \caption{Correctly assigning People to properly predicted o-spaces}
\end{figure}

\section{Experiments}
\subsection{Hyper-Parameter Tuning}
We experimented with various hyper-parameters. Our largest hyper-parameter was our model architecture. First, we had to decide on our PointNet sub-architecture. For this part, the output had to have enough dimensions to adequately capture information on all the input people 18-dimensional vectors. We experimented with output sizes of 512, 1024, and 2048 and settled on 1024 through cross validation. For our fully-connected architecture which inputs the MobileNet features and PointNet features, we experimented with various architectures and quickly found that only shallow networks worked. Our final model had two layers with the first layer having 1024 neurons. When we used deeper networks, the network was too complex and we had not enough data for our network to learn.

We also experimented with various standard deviations for our Gaussian representation of ground-truth o-spaces. Our initial Gaussian used 1 m standard deviation but we found that when two groups were close together, the deviation made it hard to distinguish the groups and our model almost invariably predicted them as one large group. We then used .1m standard deviation and this produced outputs which resembled one hot encodings. However, this caused issues because given the high dimensional output, if our network got the prediction wrong, it had trouble learning anything. This caused our network to produce very poor outputs which were always either near 1 or 0. To combat both of these issues, we settled on .5m as our standard deviation, because it allowed enough separation between groups that we could distinguish them in our prediction, but our network could still learn.

We also tuned three hyper-parameters from group assigning: maximum distance between groups, threshold for non-maximal suppression, and maximum distance to assign to a group. Maximum distance between groups determines how close we allow two potential group proposals to be. Threshold for non-maximal suppression determines how high our probability of a local maximum in our output image being a group needs to be for it to be a valid o-space proposal. Maximum distance for assigning determines how far a person's proposed o-space can be from a potential group o-space before we do not allow that person to be assigned to that group. We ran an iterative search over these three hyper-parameters in our validation set to determine the ideal combination.
\subsection{Data Imbalance}

We found that our network was doing very well on the train and validation set and outperforming existing models, so we decided to check our performance on the test set. However, upon doing so, our f1 score dropped by $30\%$. We decided to investigate why this occurred. We saw that over $75\%$ of our training and validation data contained only one group, while only $44\%$ of our test data contained one group. Since we assigned train, validation, and test data sequentially to avoid training on images identical to test images, we hypothesize that by the end of the cocktail party participants splintered into smaller groups. We also hypothesized that we had over-fit to our training and validation sets by only predicting large groups, and thus we decided to retrain our model by more heavily weighting examples with multiple groups. We present the results of both experiments.

\section{Results}

\subsection{Evaluation Metrics}

Standard accuracy evaluation metrics define a group as correctly estimated if $\ceil*{T*|G|}$ of their members are correctly estimated and if no more than $1 - \ceil{T*|G|}$ false subjects are identified, where $|G|$ is the cardinality of the labeled group $G$ and $T$ is a defined tolerance threshold. Standard values of $T$ to investigate are $2/3$ and $1$ $\cite{setti2015f}$.
It is standard to define $TP$ (true positive) to be a correctly detected group, $FN$ (false negative) to be a non-detected group, and $FP$ (false positive) to be a group that was detected but did not exist. We then measure our accuracy with three metrics: precision, recall, and $F_1$ score. Precision is $\frac{TP}{TP + FP}$, recall is $\frac{TP}{TP + FN}$ and $F_1$ score is $2*\frac{precision*recall}{precision + recall}$.

\subsection{Results}
As mentioned before, we present results from two experiments: one where all examples are weighted equally and one where we assign a proportionally higher weight to examples with multiple groups, in order to combat the data imbalance. Results for both experiments can be seen in the tables below, as well as those of the best performing model to date. Since all other prior models were purely algorithmic and not learning methods, other literature only has one accuracy to report instead of train/val/test.

\begin{figure}[h!]  
  \centering
  \includegraphics[width=0.4\textwidth]{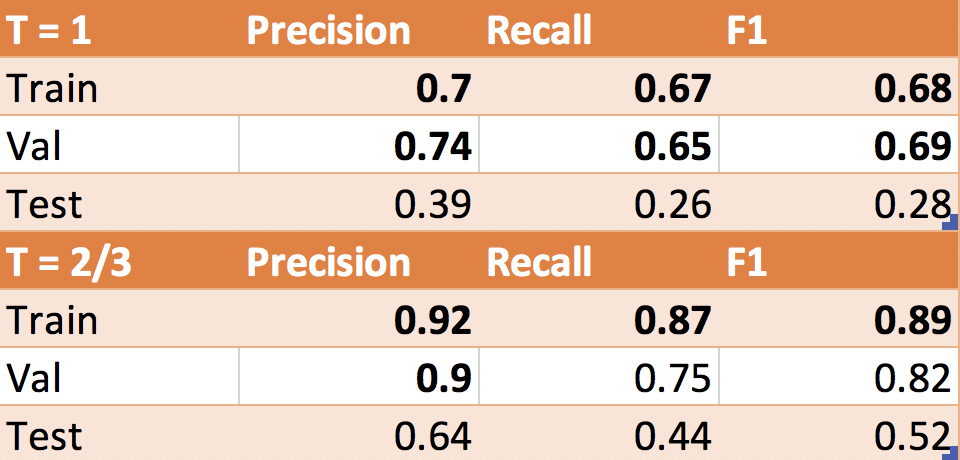}
  \centering
  \caption{Precision, Recall, F1 scores for unweighted model. Bold results outperform the best published results}
\end{figure}

\begin{figure}[h!]  
  \centering
  \includegraphics[width=0.4\textwidth]{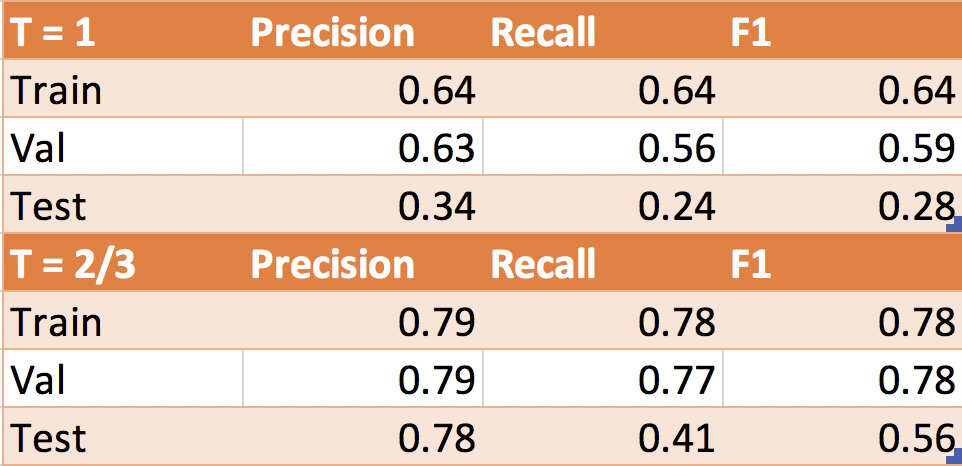}
  \centering
  \caption{Precision, Recall, F1 scores for weighted model.}
\end{figure}

\begin{figure}[h!]  
  \centering
  \includegraphics[width=0.4\textwidth]{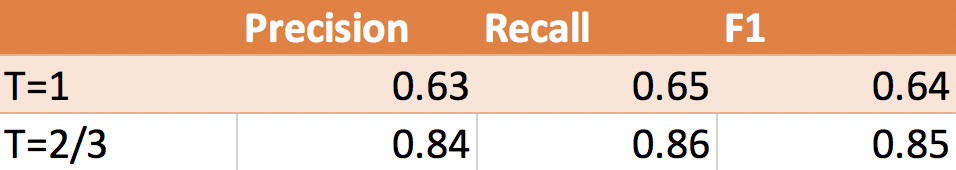}
  \caption{Best Performing Model To Date $\cite{setti2015f}$}
\end{figure}

\subsection{Discussion}
In our initial experiment with no weighting, our training and validation accuracies often outperform the best models in existence. Our $T=1$ validation $f_1$ score of $.69$ outperforms the previous best score of $.64$ which itself far outperforms all models in existence. Our $T=2/3$ validation $f_1$ of $.82$ is close to the best score of $.85$ and similar to all other scores. Our train accuracies outperform all models in existence. In fact, our train and validation accuracies nearly match accuracies when we simply assign people to the ground truth o-spaces. However, as discussed before, we experience a huge decline in performance in the test set. This is due in part to over fitting to validation and the high percentage validation examples with only one group, but also because it is simply much more difficult to predict multiple groups. For this reason it is slightly unfair to compare our test accuracy to other models which were evaluated on the entire dataset, the beginning of which was much friendlier to all algorithms. We hypothesize that if re-run on a more representable portion of our dataset, our test accuracies would improve to more standard values.

When weighting the data to attempt to balance the number of examples with multiple groups with the single-group examples, we notice a slight drop in our training and validation accuracies and a slight increase in our test accuracies. However, this increase was not enough to make our test accuracies rival best known models. The lack of improvement on the test data supports our theory that the test data is simply much harder to predict than the train/validation data and that our errors were influenced by an unlucky train/test split.

\begin{figure}[h!]  
  \centering
  \includegraphics[width=0.4\textwidth]{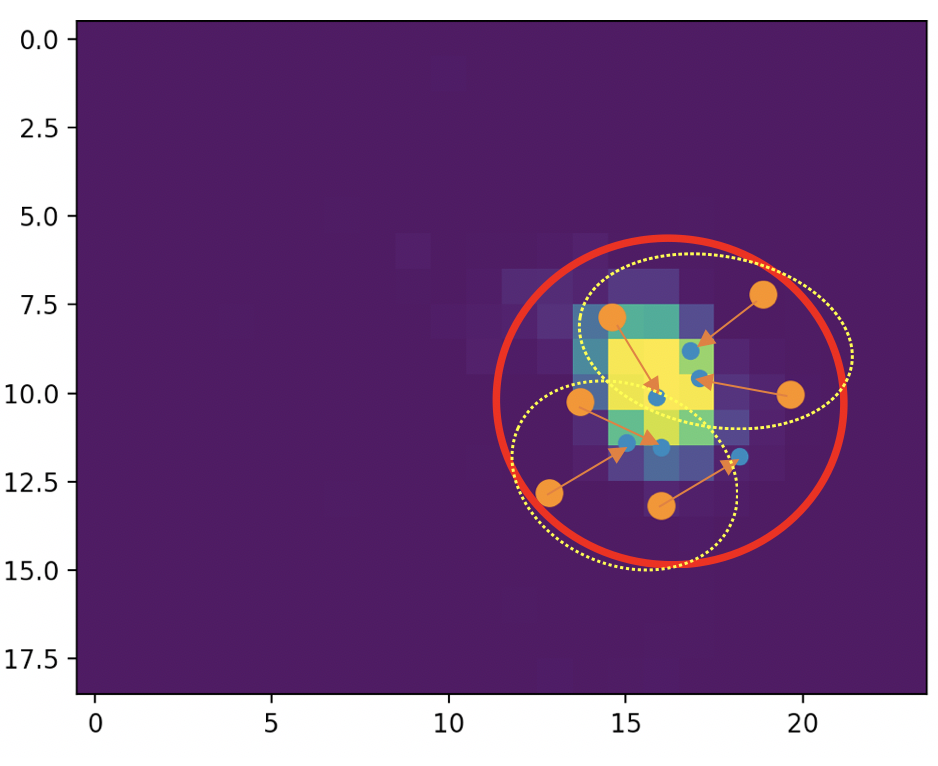}
  \centering
  \caption{Difficult to predict two nearby groups.}
\end{figure}

\begin{figure}[h!]  
  \centering
  \includegraphics[width=0.4\textwidth]{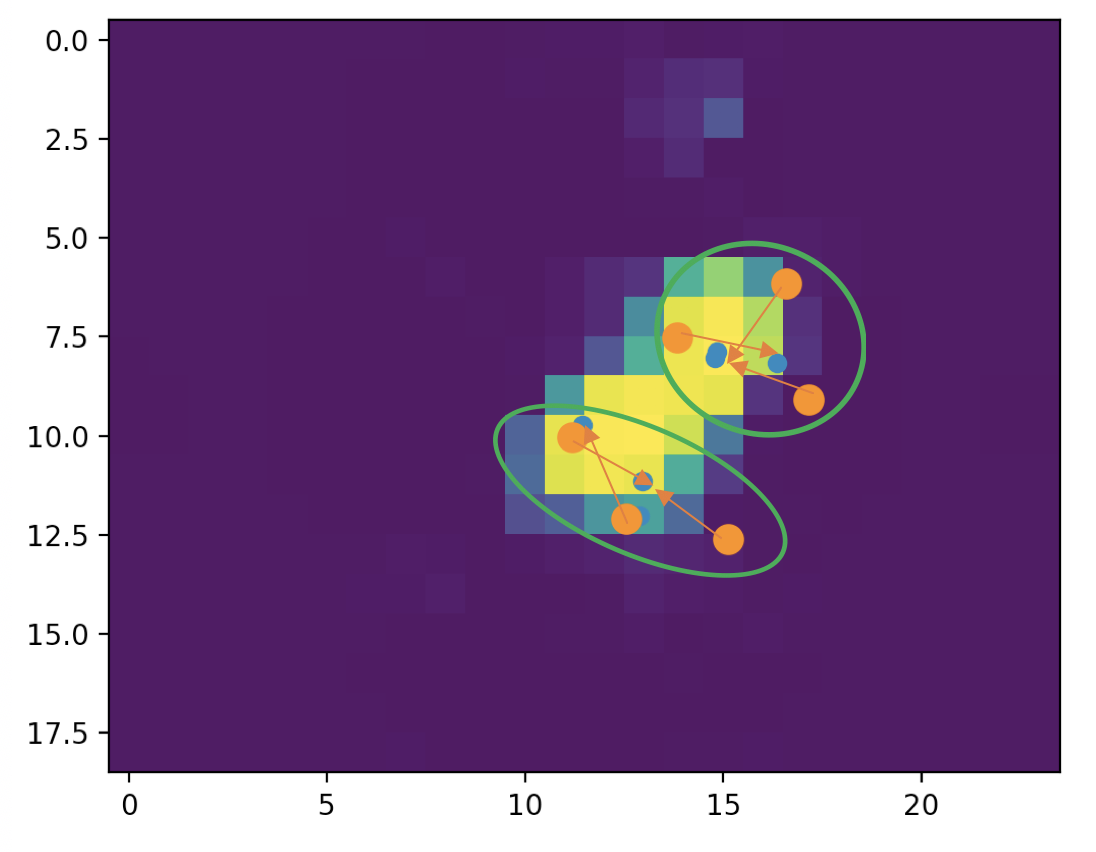}
  \centering
  \caption{Our model sometimes shows the ability to detect two nearby groups}
\end{figure}

\section{Conclusion and Future Work}
Our model showed immense promise in its ability to outperform the best models in its training and validation accuracies. However, due to a combination of over-fitting and the difficult nature of the test dataset, we failed to achieve a significant increased in performance. However, given that our model was only trained on one dataset and the extent to which our results improved when data was augmented, there is reason to believe that training on more data will allow our model to generalize and replicate its impressive training and validation performances. Thus, future development of our deep model which incorporates more examples and is not unlucky to have its test data be the most difficult subset of the data might significantly outperform existing methods. Additionally, using more datasets in varied locations will allow the extracted room features to be of more use, instead of simply indicating when we rotated the room. One downside to a convolutional or deep model as opposed to algorithmic methods proposed is that it becomes more difficult to work with varied room sizes. For different room sizes, future work will need to add padding to smaller room images so that spacial dimension are not affected. To eventually switch to a practical deep model to be incorporated in a social robot, many more datasets of different sizes must be used and the ability for variable-sized rooms must be implemented. 

{\small
\bibliographystyle{ieee}
\bibliography{egpaper_final}
}

\end{document}